# AxialGen: A Research Prototype for Automatically Generating the Axial Map[*]

Bin Jiang and Xintao Liu

Division of Geomatics, Department of Technology and Built Environment
University of Gävle, SE-801 76 Gävle, Sweden
Email: bin.jiang@hig.se, xintao.liu@hig.se

**Abstract**
AxialGen is a research prototype for automatically generating the axial map, which consists of the least number of the longest visibility lines (or axial lines) for representing individual linearly stretched parts of open space of an urban environment. Open space is the space between closed spaces such as buildings and street blocks. This paper aims to provide an accessible guide to software AxialGen, and the underlying concepts and ideas. We concentrate on the explanation and illustration of the key concept of bucket: its definition, formation and how it is used in generating the axial map.

## 1. Introduction

The generation of the axial map has been a key aspect of space syntax research (Hillier and Hanson 1984, Hillier 1996), which has conventionally relied on hand-drawn axial maps of urban environments for urban morphological analysis. Although various attempts have been made towards an automatic solution (Peponis et al. 1998, Turner et al. 2005, Batty and Rana 2004, Carvalho and Batty 2005), few of them can produce an axial map that is identical to the hand-drawn one. Instead of generating an axial map that is identical to the hand-drawn one, we focused on generating an axial map that consists of the least number of longest visibility lines that fully cover the open space of an urban environment. Open space is the space between closed spaces such as buildings and street blocks. Open space can be also called free space, implying that people can move freely around it.

We developed a two-stage solution. First, it generates all rays to entirely cover the open space of an urban environment. The rays are constructed from the vertices of the medial axes of the open space, or from the discrete points of a recursively generated ray within open space. Second, the rays are reduced to an economic set representing the axial lines. Our innovation lies in a simple yet elegant reduction algorithm based on a novel concept of bucket for representing an individual part of the open space of an urban environment.

Our solution can be justified by the notions of small scale and large scale spaces. Each axial line is a small scale space that is roughly represented by a bucket, and all those rays within the bucket are considered to be redundant and shall therefore be removed. The key to the solution lies in partitioning large scale open space into numerous small scale spaces, each of which is represented by a bucket. The bucket is not necessarily convex, so it is not a small scale space per se. However, it can roughly be considered to be a small scale space, as it can be seen entirely from a ray's point of view.

---

The remainder of this paper is structured as follows. Section 2 presents basic concepts for developing AxialGen. Section 3 concentrates on the key concept of bucket: its formation and how it is used in ray reduction. Section 4 describes in detail the functions implemented with AxialGen. Finally, section 5 presents a short summary.

## 2. Basic Concepts underlying AxialGen

The basic concepts underlying AxialGen are defined and explained in this section. Some of the concepts have also been illustrated; see Figure 1.

*Small scale space:* Small scale space is small enough to be seen entirely from any single vantage point within the space (Ittelson 1973).

*Large scale space:* Large scale space is too large to be seen from a single vantage point (Ittelson 1973).

**NOTE:** A major difference between a large scale space and a small scale space lies in whether or not it is viewable or perceivable from a single vantage point, no for a large scale space and yes for a small scale space.

*Closed spaces:* Buildings and street blocks in urban environments are referred to as closed spaces that prevent people from crossing over them. It is represented as individual polygons.

*Open space:* Open space is the space between closed spaces such as buildings and street blocks in urban environments. People can freely move around in open space, which is hence also called free space. Open space is represented as a ***complex polygon with holes*** that represent individual closed spaces, surrounded by an outmost boundary. Open space is a large scale space that is not perceivable from a single vantage point. However, while walking in open space, people can perceive individual parts or units, called isovists.

*Isovist:* It is a visible space from a single standing viewpoint (Bendikt 1979). It is NOT a small scale space per se, for isovists are not necessarily convex. However, it can approximately be considered to be a small-scale space as one is able to see it entirely from the single standing viewpoint.

*Isovist ridge:* A visually dominant direction or path of light is called an isovist ridge (Batty and Rana 2004, Carvalho and Batty 2005). In this paper it is also called a **ray**.

*Medial axes:* Medial axis is a method for representing the shape of an object by the topological skeleton - a set of curves that run along the middle of the object (Blum 1967, Blum and Nagel 1978). In this paper, it simply refers to the set of curves, hence medial axes in plural form. Every medial axes' vertex has at least two associated points from which the vertex is created, i.e., the vertex is the middle point of the associated points.

*Axial lines:* They are defined as the longest visibility lines crossing each other and running through open space of an urban environment (Hillier and Hanson 1984). The least number of axial lines constitute what we call an axial map.

*Axial map:* An axial map is defined as an economic set of axial lines, i.e., the least number of axial lines (Hillier and Hanson 1984).

***Complex polygon with holes:*** It is used to represent the open space of an urban environment. The holes are closed spaces, such as buildings and street blocks, usually surrounded by an outmost boundary.

***Voronoi regions:*** The medial axes surrounding individual closed spaces constitute the Voronoi regions of the closed spaces. In other words, medial axes are the edges of the Voronoi regions.

***Bucket:*** A bucket is defined as a space surrounding an axial line or ray, and it can be seen entirely from the point of view of the axial line or the ray (Jiang and Liu 2009). It is in this sense that a bucket can be roughly considered to be a small scale space. It is represented by a polygon surrounding an axial line or a ray. The section that follows presents more detail about the key concept underlying AxialGen.

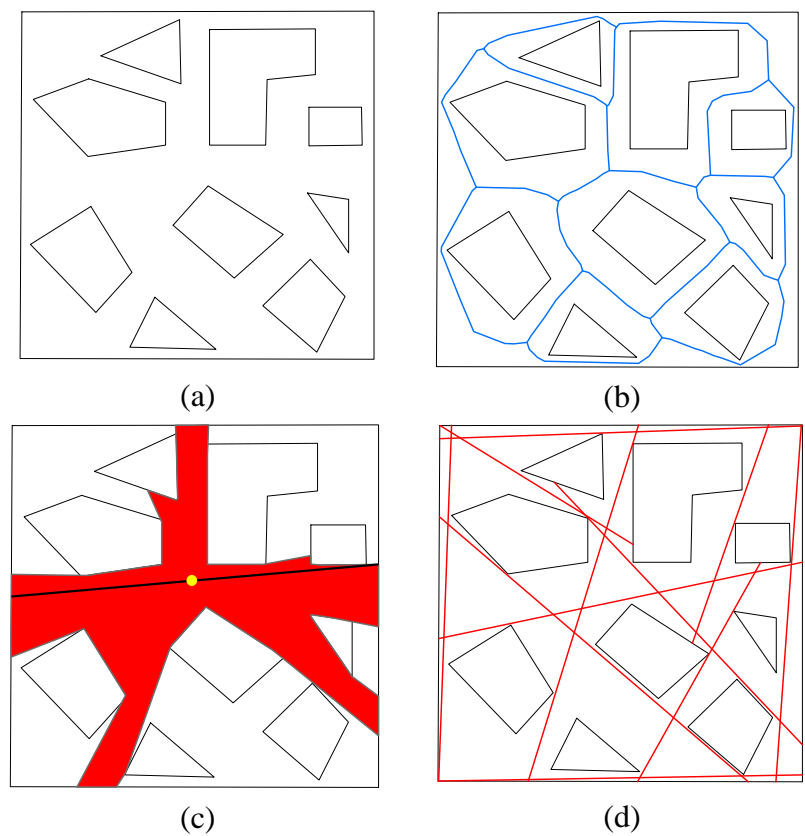

Figure 1: (color online) Illustration of various concepts: (a) open space represented by a complex polygon with holes that represent closed spaces, (b) medial axes (blue curves) of the open space constituting the edges of the Voronoi regions of the closed spaces, and (c) an isovist (red), its ridge or ray (black line above the isovist), and the standing point (yellow spot), (d) axial map that consists of the least number of the longest visibility lines (or axial lines)

## 3. Bucket Formation and Ray Reduction

The key concept underlying AxialGen is the bucket, which needs more explanation and discussion. As we can see in Figure 2, a ray cuts across four Voronoi regions of four closed spaces A, B, C and D. More specifically, the ray with two ending points, e1 and e2, cuts the Voronoi regions at vertices x1, x2 and x3. For the ending points, we can find in total four associated points, i.e., e11 and e12 for e1, e21 and e22 for e2. In addition, the ray segments x1x2 and x2x3 correspond to four medial axes segments x1y1, x2y1, x2y2, and x3y2 that intersect at points y1 and y2. The point y1 has two associated points y11 and y12, and the point y2 has two associated points y21 and y22. The two ending points, and the associated points, can be chained together to form a closed polygon - the bucket of the ray. In a more general term, the points to be chained together are e1, e2, e11, e12, e21, e22, y11, y12, . . . yn1, yn2, where n is the number of the intersection points of the corresponding medial axes segments.

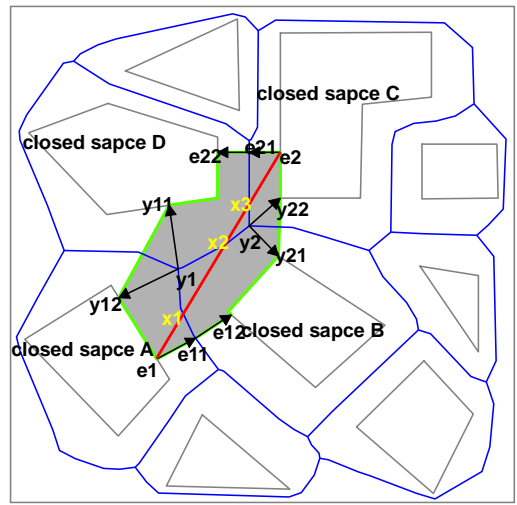

Figure 2: Illustration of the derivation of the bucket of a ray
(Note: red line = ray, blue lines = medial axes, gray lines = edges of closed spaces, arrows = association, green boundary = bucket generated for the ray)

Given that open space is fully filled with numerous rays, buckets can help to get rid of redundant rays. Eventually, the least number of the longest rays are left over to form the axial map. The reduction is done through the removal of rays within the bucket of another ray. This reduction procedure usually starts with the longest ray and goes to the next ray in a decreasing order. The resulting axial map consists of the least number of the longest axial lines. It sounds a bit surprising how the least number of the longest axial lines can be achieved. In fact, the power of the bucket concept lies in its ability to remove redundant lines. As we will see in the following section, AxialGen develops a special strategy to generate the axial map. On the one hand, we select, either globally or locally in a decreasing order, axial lines from the first longest, the second longest,…, so that all the selected lines are the longest ones. On the other hand, the bucket algorithm guarantees that no redundant or repeated axial lines exist.

## 4. Functions of AxialGen
### 4.1 Generating the axial lines
According to the definition of the axial map, we first generate all the longest visibility lines, i.e., the individual isovist ridges. There are two different ways of generating the isovist ridges: first from the vertices of the medial axes, and secondly from the discrete points of a recursively generated ray. These two ways lead to two different approaches of generating the axial lines implemented respectively with tools AxialGenG and AxialGenL. Let us take a look at the approaches or tools one by one.

The first approach takes all the vertices of the medial axes as the standing points, and generates corresponding isovist ridges, also called rays that are fully filled in open space of an urban environment. This is the first stage. Then in the second stage we reduce the rays into an economic set representing the axial lines that constitute an axial map. For the reduction procedure, we take two different strategies. The first strategy works like this. It selects the first longest line (or ray), and subsequently removes all those rays 85% overlapped with the bucket of the selected line, then the second longest line and removing all those rays 85% overlapped with the bucket of the second selected line, ... till there is no ray left in the ray set. It is a loop procedure. The second strategy selects the first longest line, and removes all those rays 85% overlapped with the bucket of the first selected line by sorting all the rays. Other longest lines are obtained, not by sorting all the remaining rays, but by sorting those that intersect with the first selected line. The second strategy works in a recursive fashion. It can be further divided into two different ways with respect to two different tree searches: Breadth First Search (BFS), and Depth First Search (DFS). Comparing the two tree searches, we found that BFS tend to be the better one. This is because that with BFS, we can add both line length and line connectivity as selecting criteria.

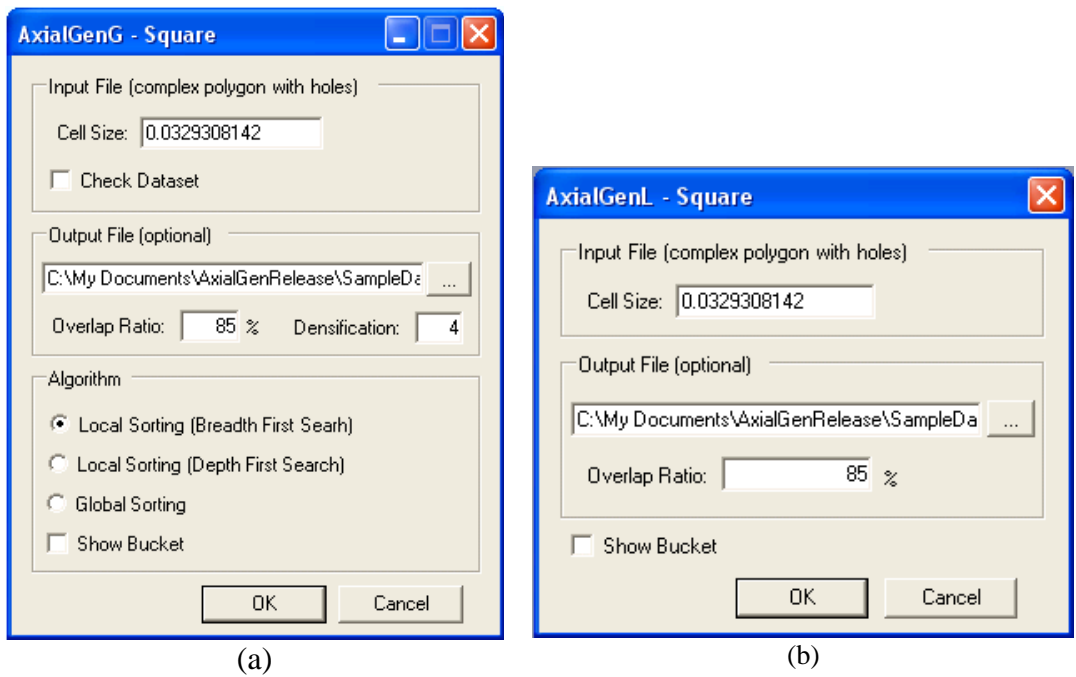

Figure 3: Parameter settings for (a) AxialGenG, and (b) AxialGenL

It is worth while to add a few words on the parameters shown in Figure 3: cell size, overlap ratio, and densification. Cell size is the resolution for generating medial axes and it is determined by one third of the shortest distance between the two closest closed spaces or between a closed space and the outmost boundary. Removing lines within a bucket is just an idealized case (100% overlapped), so in fact we adopt an overlap threshold to determine whether or not a line is redundant. The overlap ratio is set at 85%, implying that a ray should be removed if 85% of it is within the bucket of another ray. In some cases we need to densify the vertices of medial axes, and this particularly occurs to some parts that have very regular shapes, e.g., a T-shaped space. It is considered to be too scattered, if the distance between two adjacent vertices is more than four times that of the average width of medial axes, i.e., the sum of the distances from a vertex to its associated points on the edges of closed spaces or the outmost boundary.

The second approach starts with an arbitrarily drawn line in the open space of an urban environment. This line is stretched in two directions to hit the edges of the closed spaces or the outmost boundary, and it becomes the first ray. This first ray is discretized into individual points from which we create isovist ridges or rays. Those rays with 85% within the buckets of rays will be removed based on our bucket reduction algorithm. This procedure goes on until the entire open space is covered with the least number of the axial lines. With this approach, there is a resolution issue, i.e., how many discrete points should be adopted for generating the rays? We adopt one third of the shortest distance between two closest closed spaces or between a closed space and the outmost boundary.

**4.2 Accessories with AxialGen**
There are three accessories namely MedialAxesGen, IsovistExplorer, and BucketExplorer. MedialAxesGen is used for generating the medial axes of a complex polygon with holes, and it is an essential part for generating the axial lines. We package it as its own tool for the sake of convenience. Another independent tool is IsovistExplorer, which is also an essential part of AxialGen, used for creating isovist ridges or rays. With this tool, the user can interactively explore the isovist patterns and the isovist ridges with a complex polygon. The most useful accessory tool is probably BucketExplorer, with which the users can interactively explore the bucket of an arbitrarily drawn ray, or that of axial lines. It is particularly useful for checking whether or not there are redundant axial lines, i.e., for simply checking if an axial line is within another axial line's bucket.

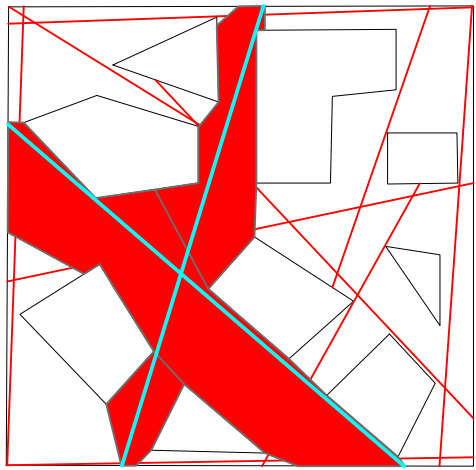

Figure 4: An example showing overlapped buckets with BucketExplorer

**5. Summary**
This short paper aims to provide an accessible guide to the concepts and ideas underlying the AxialGen software program for automatically generating the axial lines. We explained and illustrated some important concepts including open space versus closed spaces, large-scale space versus small-scale space, isovists versus isovist ridges (or rays), medial axes versus axial lines, axial maps, and buckets. We particularly focused on the key concept of bucket we developed for reducing numerous rays into an economic set of the axial lines, i.e, the least number of the longest visibility lines that constitute the axial map. For a more comprehensive introduction to the algorithms, the reader is encouraged to refer to Jiang and Liu (2009). To try the software, please download it from http://fromto.hig.se/~bjg/AxialGen/, and install the client edition in your machine, or simply use the more powerful server edition. We have yet to assess practical usability of the software in various applications.

**Acknowledgment**
We thank Petra Norlund for polishing up our English

**Appendix: Installing AxialGen 1.0 Extension for ArcGIS 9.2**

(1) Check for Microsoft .NET Framework 2.0
Before you install AxialGen 1.0, make sure that Microsoft .NET Framework 2.0 has been installed.

You can download the .NET Framework 2.0 from the Microsoft Download Center. If you have installed this successfully on your machine, you should be able to identify this from the **Add/Move Programs** list in the Control Panel of Microsoft Windows.

(2) Installing AxialGen is a pretty straightforward process. You can download the AxialGen package from this website (http://fromto.hig.se/~bjg/AxialGen), and save it to a location on the computer. Extract the installer file and double click the **AxialGenClientInstall.msi** to launch setup:

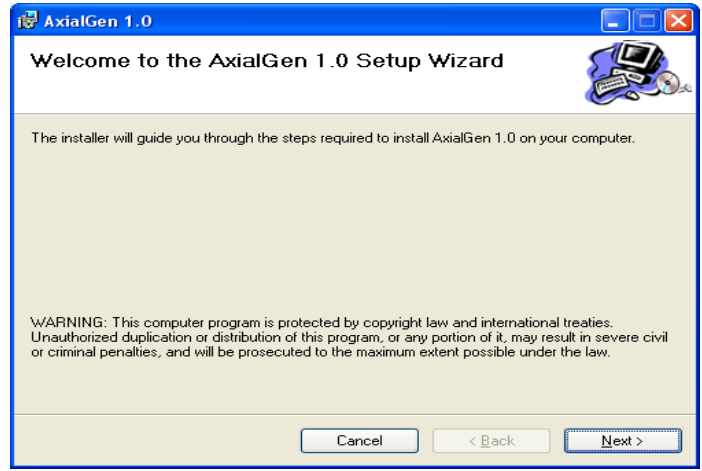

Click **Next**. The screen for specifying the installation location will appear as follows:

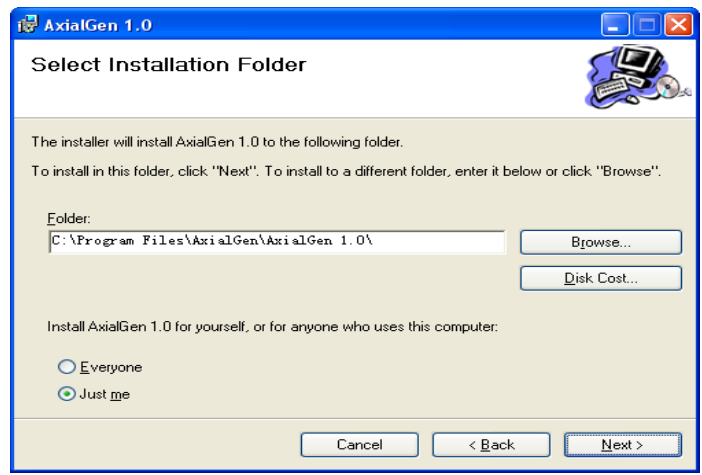

You may accept the default installation path, or change it as you wish.

To activate the extension upon the completion of the installation, follow the subsequent steps:
1) Click the **Start** menu, point to **All Programs** and Start **ArcMap**
2) Click on ArcMap's **Tools** Menu and then choose **Extension.**
3) Check **AxialGen 1.0** as follows.

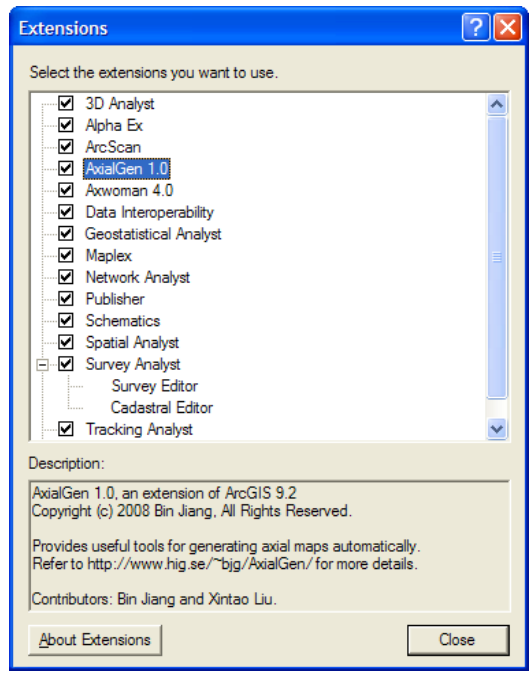

4) Click **Close**.
5) Click **View**, point to **Toolbars**, and select AxialGen 1.0.

AxialGen 1.0 toolbar has five buttons:

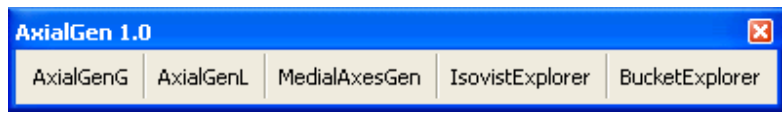